\documentclass{article}
\usepackage{spconf,amsmath,graphicx}

\usepackage{float}
\usepackage{amsmath,graphicx,amsfonts,amssymb,url}
\usepackage{setspace}
\usepackage[caption=false]{subfig}
\usepackage{blindtext}
\usepackage[inline]{enumitem}
\usepackage{color}
 \usepackage{subfig}
\usepackage{multirow}
\newsavebox{\measurebox}
\usepackage{minibox}
\usepackage{array,multirow,booktabs}
\usepackage{graphicx}
\usepackage{siunitx}
\usepackage{amsfonts,amssymb}
\usepackage{balance}
\usepackage{pifont} 
\newcommand{\squeezeup}{\vspace{-2.5mm}}

\usepackage{algorithm}
\usepackage{algorithmic}
\usepackage{url}

\newcommand*{\affaddr}[1]{#1} 
\newcommand*{\affmark}[1][*]{\textsuperscript{#1}}

\usepackage{color}
\newcommand{\suiyi}[1]{\textcolor{black}{#1}}

\title{Subjective and Objective Quality Assessment of Mobile Gaming Video}
\name{Shaoguo Wen\affmark[2]\thanks{Suiyi Ling and Shaoguan Wen make equal contribution, Junle Wang is the corresponding author.}, Suiyi Ling\affmark[1] , Junle Wang\affmark[2], Ximing Chen\affmark[2], Lizhi Fang\affmark[2], Yanqing Jing\affmark[2], Patrick Le Callet\affmark[1]  }
\address{\affaddr{\affmark[1] LS2N, \ University of Nantes} \ \  \affaddr{\affmark[2]Turing Lab, \  Tencent }}
 
\begin{document}
 
\maketitle

\begin{abstract}
Nowadays, with the vigorous expansion and development of gaming video streaming techniques and services, the expectation of users, especially the mobile phone users, for higher quality of experience is also growing swiftly. As most of the existing research focuses on traditional video streaming, there is a clear lack of both subjective study and objective quality models that are tailored for quality assessment of mobile gaming content. To this end, in this study, we first present a brand new Tencent Gaming Video dataset containing 1293 mobile gaming sequences encoded with three different codecs. Second, we propose an objective quality framework, namely Efficient  hard-RAnk  Quality  Estimator  (ERAQUE), that is equipped with (1) a novel hard pairwise ranking loss, which forces the model to put more emphasis on differentiating similar pairs; (2) an adapted model distillation strategy, which could be utilized to compress the proposed model efficiently without causing significant performance drop. Extensive experiments demonstrate the efficiency and robustness of our model.

\end{abstract}
\section{Introduction}
\label{sec:intro}
Gaming video streaming is composed of a wide-range of online services including gaming video streaming, esports broadcasting and cloud gaming
services. In the past decade, the booming international popular gaming video streaming platforms including Twitch, YouTube, Smashcast, Afreeca TV, Gosu Gamers, \textsl{etc.}, and cloud gaming platforms like Google Stadia and Nvidia Geforce Now, are gradually taking a large share of video streaming~\cite{barman2018gamingvideoset}. Most of these services have also mobile app to meet the growing prevalence of mobile usage in modern life. According to~\cite{ling2020towards}, gaming content also makes up a significant proportion of of User Generated Contents (UGC) on social media platforms, where a certain amount of users browse/view the contents using their mobile phone by a daily base. As reported in~\cite{game_trend}, mobile gaming accounted for 58.8\% of the digital games market in 2019. This percentage is growing even more dramatically due to the global COVID-19 pandemic in 2020. Users' higher requirements for better Quality of Experience (QoE) necessitate robust quality control of these gaming contents. 
Different from the contents on common video streaming platforms~\cite{ling2020towards}, quality assessment of gaming videos raise new challenges to the community as: their temporal complexity could be significantly higher; unlike natural videos, they are graphical rendered/generated contents; most of the gaming streaming platform requires real-time quality evaluation; and cloud gaming is sensitive to delay. Notwithstanding the fact that subjective study is time-consuming and expensive to conduct, it is still the foundation for the development of rigorous objective quality metrics. As most of the existing subjective studies focus mainly on video on traditional Video on Demand streaming services~\cite{barman2018gamingvideoset}, studies conducted for gaming contents, especially for mobile applications, are still scarce.  

With the burgeon of video encoding techniques and hardware based acceleration frameworks, dedicated fast codecs have been developed to alleviate relevant stresses. The development of the video objective quality metrics should also keep pace with the one of the codecs. Among No-Reference (NR), Reduced-Reference (RR), and Full-Reference (FR) metrics, NR metrics are of greater value, and significantly more piratical for real-time gaming content quality evaluation~\cite{zadtootaghaj2020demi} since pristine reference is not always existing or accessible. Advanced deep learning based NR quality metric enjoyed a big leap in performance in recent years. The development of Deep Learning (DL) has brought along a new wave of NR quality assessment models that perform markedly better than traditional metrics. However, very few of them target the gaming application, typically for mobile users. Recall that most of those DL based models are complex, cumbersome and thus barely could be deployed on mobile directly for real-time prediction.  
 
Based on the discussions above, in this study, we present a novel large-scale mobile gaming video data set, and a brand new \suiyi{Efficient hard-RAnk Quality Estimator (ERAQUE)} to remedy the lack of existing subjective and objective studies.

 \squeezeup  \squeezeup
\section{Related Work}
\label{sec:RW}\squeezeup
\textbf{\ \ \ \ \  Subjective study:}  Recently, an image dataset for gaming content was released~\cite{ling2020subjective}, where each individual image was labeled with four dimensions including the `overall quality' dimension. However, in this study the `overall quality' was subjectively assessed from an aesthetic perspective of view. The `GamingVideoSET'~\cite{barman2018gamingvideoset}, consists of 576 sequences encoded using H.264 from 24 gaming content. KUGVD~\cite{barman2019no} is another public available gaming video dataset, but only 6 source contents were collected. 
Cloud Gaming Video Dataset (CGVDS)~\cite{zadtootaghaj2020quality} is one of the largest existing public gaming video dataset in the domain. Although the number of considered contents in this subjective study is slightly larger than the one in GamingVideoSET, more training data is still required for mainstream deep learning models. Most importantly, none of the above subjective studies were conducted for mobile games, with mobile phones.

\textbf{Objective models:}
The demand of No-Reference quality metrics is growing significantly, especially for real-time quality control. Several dedicated blind quality metrics were developed to meet the rapidly growing need. In~\cite{utke2020towards}, Utke \textsl{et al.} explored different Convolutional Neural Network (CNN) architectures, \textsl{e.g.,} DenseNet$_{121}$, without the help of reference video, to predict VMAF score for gaming videos. A blind quality metric was presented in~\cite{zadtootaghaj2018nr}, where Support Vector Regression (SVR) was applied to regress quality scores using frame-level index. Recently, the G.1072 planning model~\cite{g1072}, which evalutes the quality of gaming videos based on the videos' information, including metadata, codec information, etc., was summarized in the ITU-T recommendation standard (ITU-T.G1072).  Similarly, a parametric video quality metric was introudced in~\cite{zadtootaghajMMSys} by Zadtootaghaj~\textsl{et al.} Among existing NR video metrics, DEMI~\cite{zadtootaghaj2020demi} is one of the most recent state-of-the-art video quality gaming metric. It was developed by training a CNN to capture videos distortions guided by full reference quality metric, finetuning the obtained network on a smaller quality datasets, and finally predicting the quality scores utilizing Random Forest. Following a similar recipe, a three-step framework NDNetGaming~\cite{NDNetgaming} was designed using also CNN, with a novel temporal pooling methodology that considers the temporal masking effect. However, none of them were designed and tested on mobile gaming contents.

\squeezeup   
\section{The Subjective Study}
\label{sec:Sub} \squeezeup
In this section, details of our novel Tencent Gaming Video (TGV) dataset and the subjective experiment are provided. Overall information of the dataset is summarized in Table~\ref{tab:dataset}.

\begin{table}[!htbp]
\centering
\squeezeup  \squeezeup
\caption{Summary of out TGV dataset.}
\begin{tabular}{|c|c|}  \hline
 \textbf{Duration} & 5 sec \\  \hline 
 \textbf{Frame rate}  & 30 f/s\\  \hline 
 \textbf{Resolution}&   480P, 720P, 1080P \\\hline 
\textbf{Bitrate}  &   $100,200, ... , 10^4$  \\\hline
\end{tabular}
 \squeezeup
\label{tab:dataset}
\end{table}

\begin{figure}[!htbp]
\centering
 \mbox{ \parbox{1\textwidth}{
  \begin{minipage}[b]{0.33\columnwidth}
   \subfloat[ ]
  {\label{fig:figB}\includegraphics[width=\textwidth]{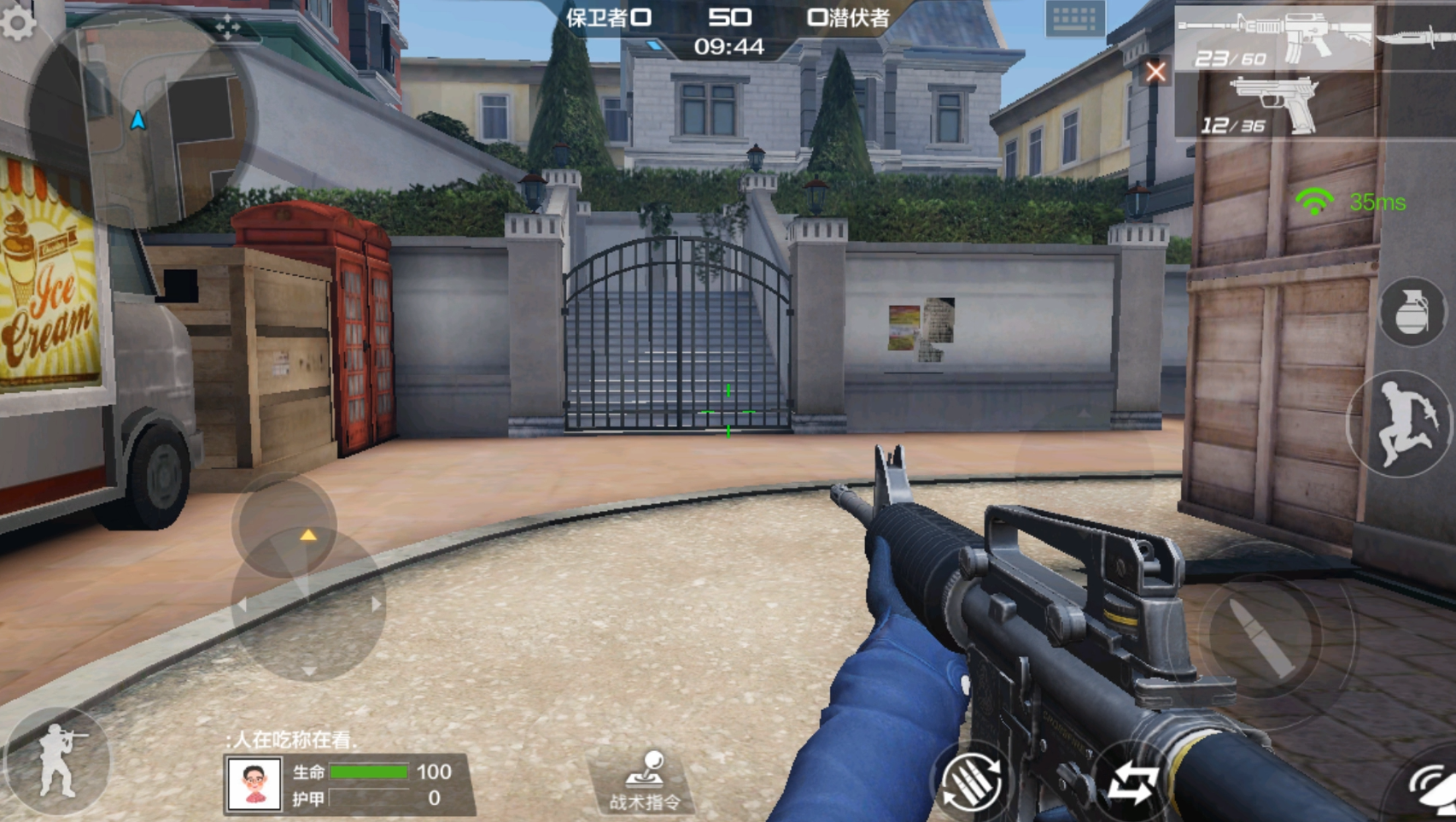}} 
  \end{minipage}
    \begin{minipage}[b]{0.33\columnwidth}
  \subfloat[ ]
  {\label{fig:figA}\includegraphics[width=\textwidth]{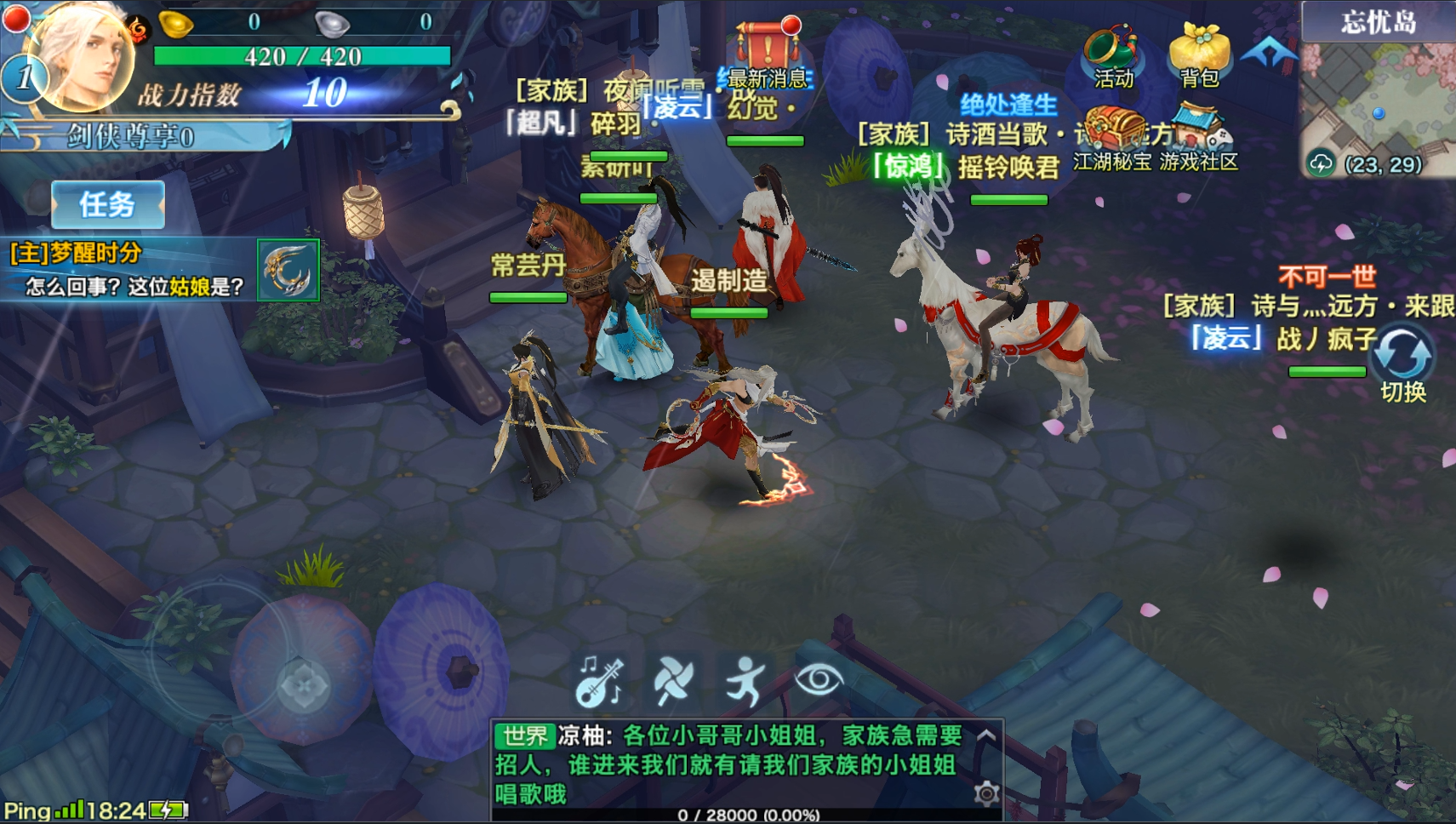}}
  \end{minipage}
    \begin{minipage}[b]{0.33\columnwidth}
    \subfloat[   ]
  {\label{fig:figC}\includegraphics[width=\textwidth]{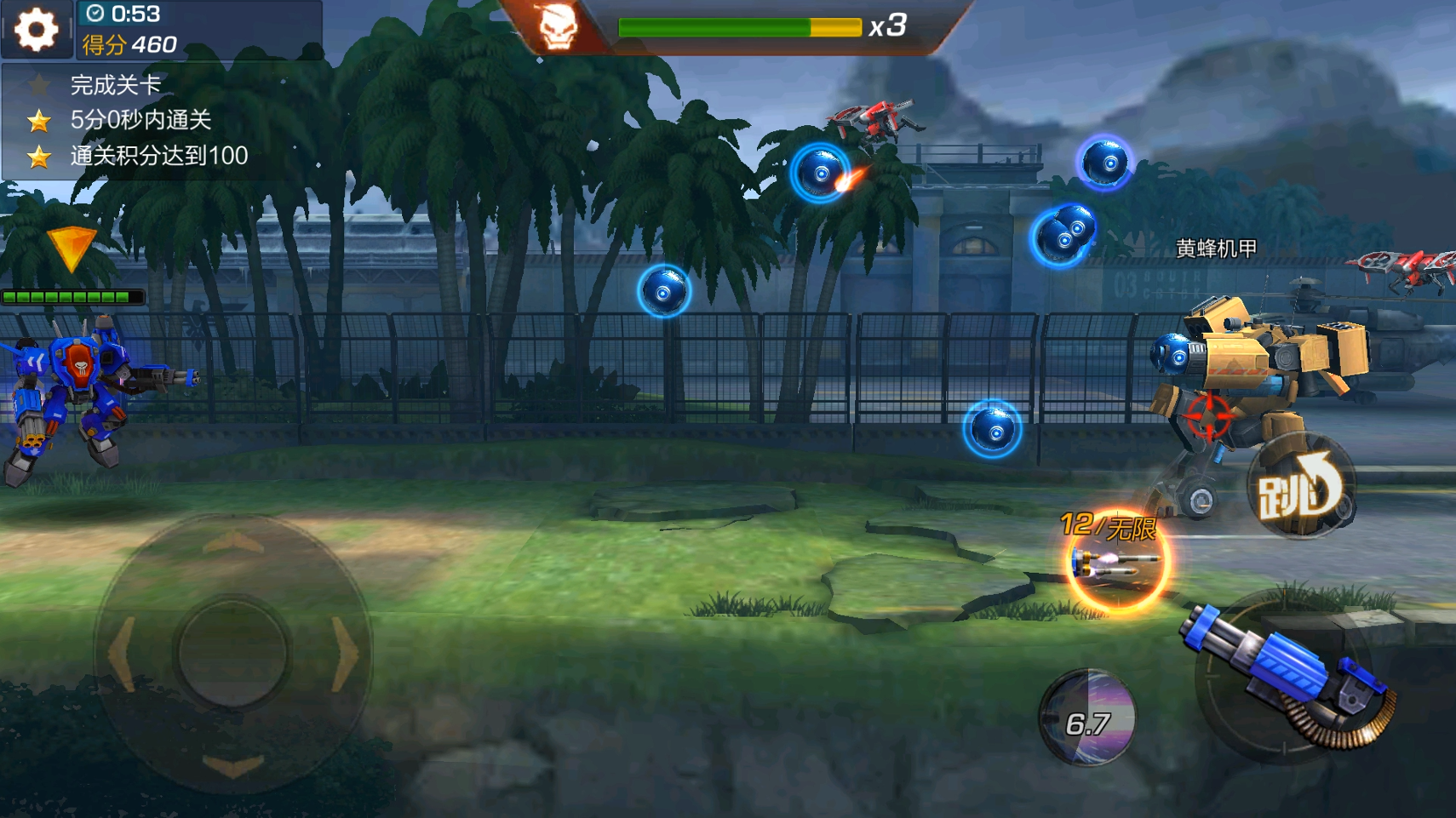}}
   \end{minipage} 
   \\
      \begin{minipage}[b]{0.33\columnwidth}
   \subfloat[ ]
  {\label{fig:figB}\includegraphics[width=\textwidth]{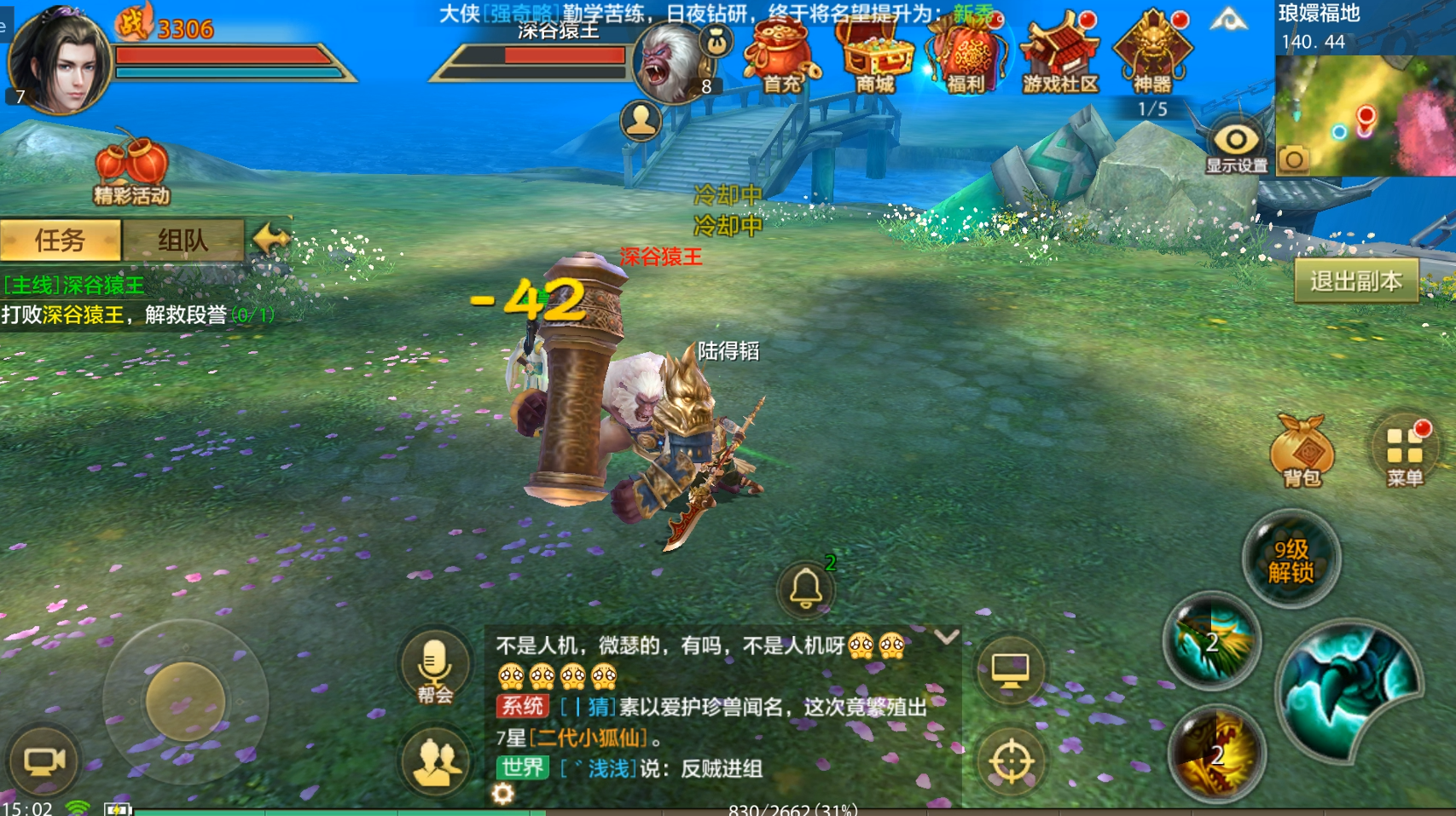}} 
  \end{minipage}
    \begin{minipage}[b]{0.33\columnwidth}
  \subfloat[ ]
  {\label{fig:figA}\includegraphics[width=\textwidth]{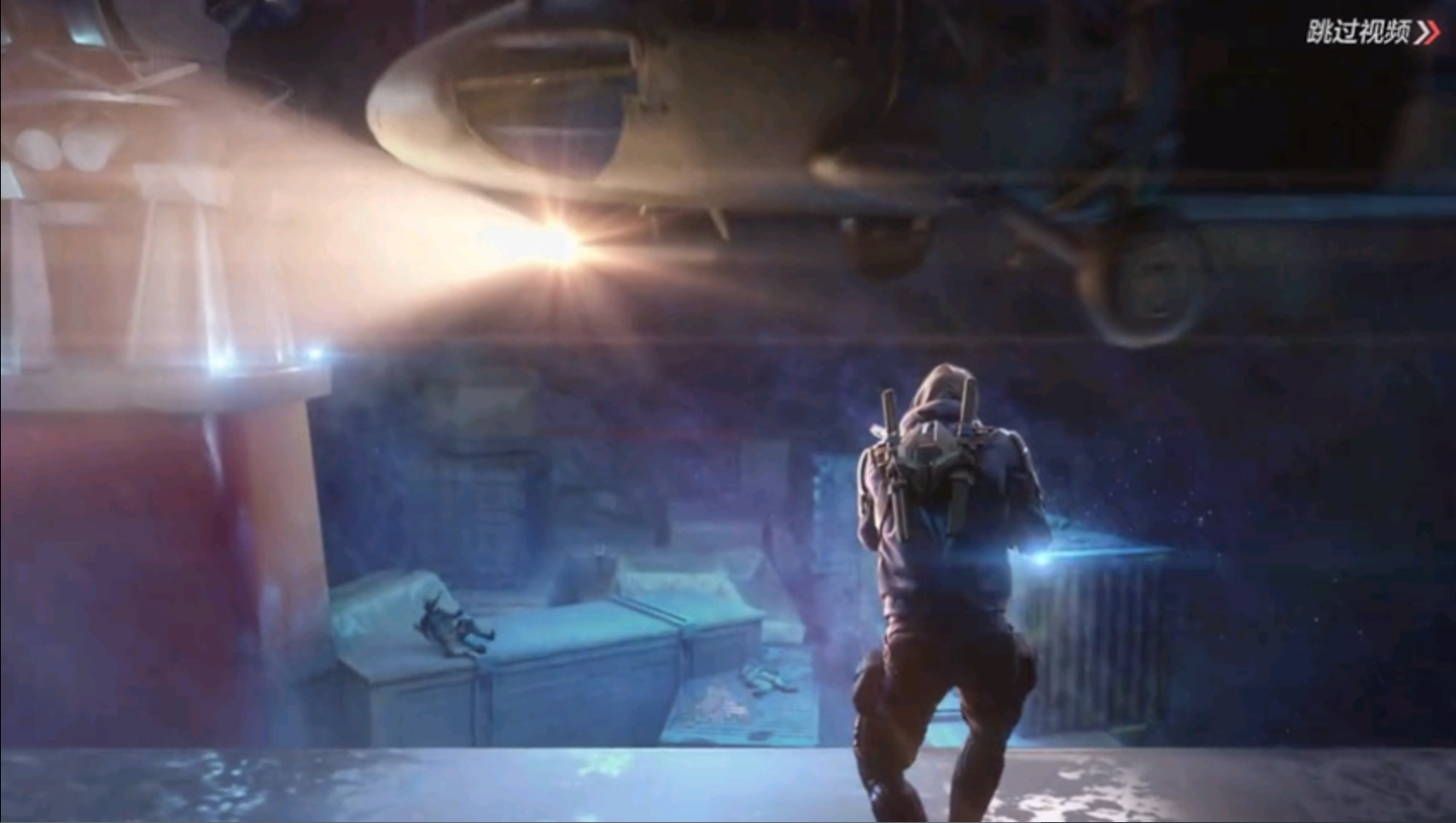}}
  \end{minipage}
    \begin{minipage}[b]{0.33\columnwidth}
    \subfloat[   ]
  {\label{fig:figC}\includegraphics[width=\textwidth]{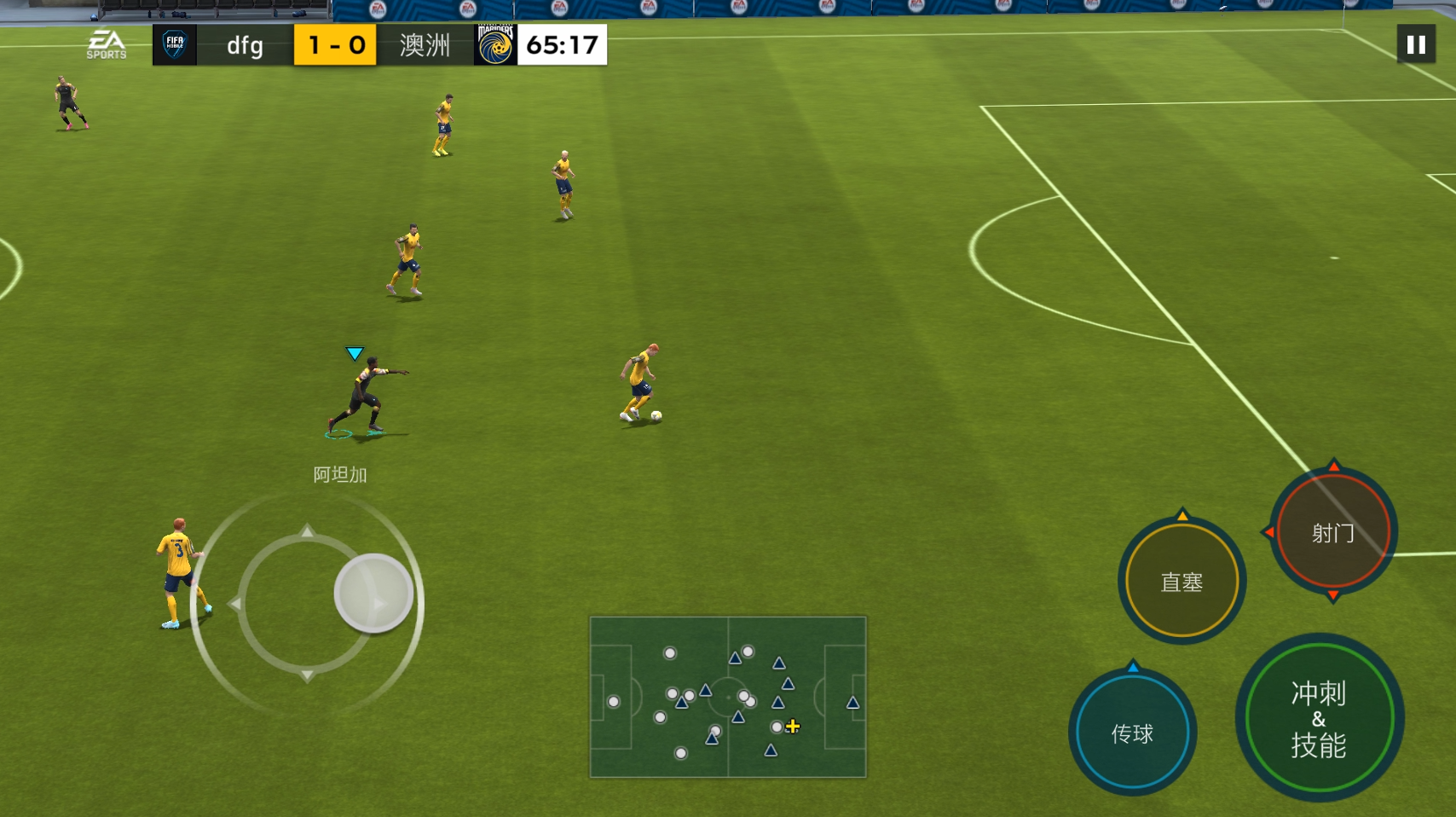}}
   \end{minipage} 
}} \squeezeup
\caption{Thumbnails of selected mobile gaming contents.}
\squeezeup \squeezeup
\label{fig:thub}
\end{figure}

\textbf{Content:} Our gaming video dataset contains 1293 sequences. In total 150 source gaming videos were collected from 17 mobile games. These games are available from the Tencent internal gaming center. Thumbnails of selected mobile gaming contents are shown in Fig.~\ref{fig:thub}. There different codecs, including the H264, H265, and one internal Tencent codec, were employed to encode each of the content with a variety of different setting/configurations, \suiyi{\textsl{e.g.,} CRF values were selected randomly from $[1, ... ,50]$.}

\textbf{Environment \& Observers:} The Absolute Category Rating (ACR) subjective protocol was used for the test. During the experiment, 19 participants with normal or corrected-to-normal acuity were asked to score the gaming videos. Among them, there were 13 males, and 6 females, most of them are between 23-26 years old. The subjective experiment was conducted in an regularized office room with, where illuminance, \textsl{etc.} was controlled according to~\cite{series2012methodology}. During the subjective experiment, the observers was asked to launch a internally developed platform to start the test, using their own mobile phone. In another word, different models of mobile phone were utilized by different participants to conduct the subjective test, which is consistent with real application scenario. After collecting the subjective data, two outliers were removed using the screening tools recommended by~\cite{series2012methodology}.

\squeezeup
\section{The Proposed Objective Model}
\label{sec:Obj}

\subsection{Network Architecture} 
\begin{figure}[!htbp]
    \centering
    \includegraphics[width=\columnwidth]{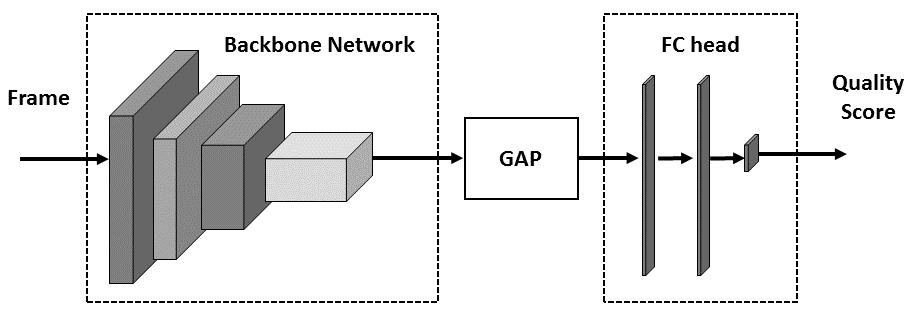}
    \caption{ Architecture of the proposed quality metric.}
     \squeezeup  
    \label{fig:arch}
\end{figure}

The architecture of the proposed \suiyi{ERAQUE for gaming video quality assessment} is depicted in Fig.~\ref{fig:arch}. In quality domain, due to limited training data, deep complex models are prone to overfitting~\cite{ling2020re}. \suiyi{Therefore, the first part of the network is a light-weight backbone network. Subsequently, the Global Averaged Pooling~\cite{szegedy2015going} is plugged in to obtain more streamlined latent representation following with two Fully Connected (FC) layers to output the final quality score. When predicting quality score for gaming videos, frames are first extracted and fed into the model, and then the frame-wise scores are averaged to obtain the final quality score of the video. Based on different use case under different scenarios, network like Mobilenet~\cite{howard2017mobilenets}, Shufflenet~\cite{zhang2018shufflenet}, ResNet-18~\textsl{etc.}, could be employed accordingly. In this study, we start with the classical ResNet 18 network that was pre-trained on Imagenet as backbone, and finetune it using quality data. This model is further compressed our re-adapted model distillation strategy. Details are shown in Section~\ref{sec:md}. }

\begin{figure}[!htbp]
    \centering
    \includegraphics[width=\columnwidth]{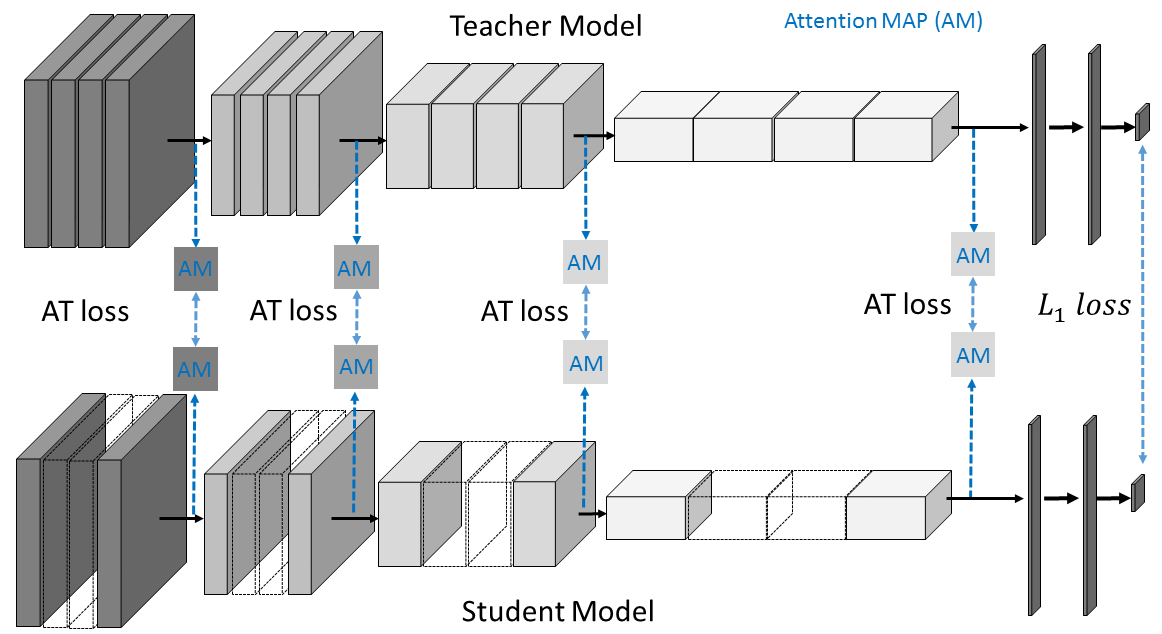}
    \caption{ Teacher-student network for model distillation. }
     \squeezeup  \squeezeup
    \label{fig:dis}
\end{figure}  
 \squeezeup
\subsection{Loss function} 
Let $\hat{y}$ and $y$ be the quality score predicted by the objective model, and the ground truth quality score collected from the subjective experiment respectively. $n$ is the total number of considered videos. 
The loss function of our proposed model is composed of two parts as defined below, where $ \lambda $ is a parameter that balances the two losses:
\begin{equation}
L = L_{mae} + \lambda \cdot L_{hard-rank}.
\label{eq:overall}
\end{equation}

The first part is the Mean Absolute Error (MAE) loss $L_{mae}$ between the ground truth and predicted scores:
\begin{equation}
L_{mae} = \frac{\sum^n_{i=1} | \hat{y_i} - y_i |}{n}.
\end{equation}
The second part is a hard pair-wise ranking loss $L^{ij}_{hard-rank}$ that is inspired by the metric learning based framework proposed in~\cite{liu2017rankiqa}. Given a pair of stimuli $(y_i,y_j)$, the proposed $L^{ij}_{hard-rank}$ is designed as:
\begin{equation}
L^{ij}_{hard-rank} = l_2(y_i, y_j) \cdot max (0, |y_i-y_j|-l_1(y_i, y_j)\cdot (\hat{y_i} -  \hat{y_j} )   ),
\end{equation}

where $l_1$ and $l_2$ are further defined in equation (\ref{eq:l1}) and (\ref{eq:l2}) correspondingly, and $\tau$ is a margin (similar to the ones defined in~\cite{liu2017rankiqa}):
\begin{equation}
l_1= \left\{ \begin{array}{rcl}
 1, \ \ \ \ \ \  y_i \geq y_j  \\  
 -1,  \ \ \  \ \ \  otherwise \ \ \ 
\end{array}\right.
\label{eq:l1}
\end{equation}

\begin{equation}
l_2= \left\{ \begin{array}{rcl}
 1, \ \ \ 0 < |y_i - y_j|\leq \tau \\  
0,  \ \ \ \ \ \  otherwise
\end{array}\right.
\label{eq:l2}
\end{equation}

In the quality domain, it is easier for most of the objective models to predict the quality scores for pairs that are of significantly different quality~\cite{ling2020strategy,li2018hybrid}. If the visual difference of two considered stimuli is high enough for observers to differentiate them, the obtained subjective data would be of less bias and errors~\cite{li2020gpm}, and thus cause less uncertainty for the prediction of objective quality models~\cite{ling2020rate}. Therefore, following a similar concept of hard sample defined in~\cite{ling2020few}, for quality assessment of gaming video, we consider the pairs that have similar quality scores as hard samples. Intuitively, $l_2$ is a gating function, where only hard sample pairs that have close quality scores ($\textsl{i.e.,} 0 < |y_i - y_j|\leq \tau$) are taken into account, and $L^{ij}_{hard-rank}$ is thus named as the `hard pair-wise ranking loss'. In this study, we first normalize the ground truth quality scores into a range of $[0,1]$, and set $\tau=0.1$ empirically.

  \squeezeup
\subsection{Model Distillation}
\label{sec:md}
In general, \suiyi{when using ResNet-like models following a de-facto standard, the network is commonly constructed by stacking a certain number  of blocks.} \suiyi{These models could be divided into 4 stages, where each} stage contains a certain number of repeated residual blocks \suiyi{following by a sequence of convolution layers that downsample the input feature}. For ResNet18, \suiyi{there are 2 residual block at each stage. Obviously, the more blocks there is, the more sophisticated the model is, \textsl{i.e.}, more parameters.} 
Knowledge distillation~\cite{hinton2015distilling,zagoruyko2016paying} can transfer knowledge from \suiyi{a complex model, \textsl{i.e.}, the teacher model, to a lighter version of itself, \textsl{i.e.}, the student model. This `teacher-student' model compression technique, is of great potential to obtain lighter network with more stationary performance, as training a small model from the scratch often ends up encountering issues like underfitting. }
 
\suiyi{The concise overview of our re-adapted teacher-student training framework for model distillation is depicted in Fig.~\ref{fig:dis}. In the framework of the proposed objective model, the backbone ResNet18 is utilized as the teacher model within the knowledge distillation pipeline. In concrete words, the backbone ResNet18 was first trained until the performance no longer increases, and was freezed during the distillation procedure, as shown in the upper part of Fig.~\ref{fig:dis}.}

\suiyi{Afterwards, ResNet18-tiny was used as the student model, where there was only 1 block at each stage and the remaining architecture is the same as original ResNet18, as presented in the lower part of Fig.~\ref{fig:dis}. Unlike the default setting considered in common classification task, essentially, our model is a regression model. Therefore, the knowledge distillation loss proposed in~\cite{hinton2015distilling} is not applicable in our use case. Hence, the Knowledge distillation (KD) loss $l_{kd}$ is re-adapted as: }  \squeezeup   
\begin{equation}
l_{kd} = |\hat{y}_{s}-\hat{y}_{t}|, 
\end{equation}
where $\hat{y}_{s}$ and $\hat{y}_{t}$ are \suiyi{the} predicted scores from student model and teacher model respectively.

\suiyi{It is worth noting that, throughout the knowledge distillation process, the student network does not only learn how to predict better the quality score, but also the latent representation using the intermediate convolution layers. Inspired by the attention mechanism proposed in~\cite{zagoruyko2016paying}, their attention loss is re-adapted to distillate intermediate feature maps between the student model and teacher model. Let $\textbf{\{A\}}$ be the set of feature map output from the aforementioned 4 stages in the ResNet network, and $A^i$ denotes feature map output from the $i_{th}$ stage. Then, we have:} 
 \squeezeup  \squeezeup
\begin{equation}
\widetilde{A}^i = \sum_{i=1}^{C^i}|A^i|,
\end{equation}
\suiyi{where} $C^i$ is the \suiyi{$i_{th}$} channel \suiyi{of} $A^i$. \suiyi{The attention loss used for transfer attention could be then defined as}:
\begin{equation}
l_{at}^i = \frac{1}{2n^i}\cdot \left\lVert \frac{vec(\widetilde{A}^i_{stu})}{||vec(\widetilde{A}^i_{stu})||_2}-\frac{vec(\widetilde{A}^i_{tea})}{||vec(\widetilde{A}^i_{tea})||_2} \right\rVert_2  ,
\end{equation}
\suiyi{w}here $vec(\cdot)$ is \suiyi{the} vectorization operation, $n^i$ represents the total number of nodes in $A^i$. Finally, \suiyi{the total loss $L_{T-S}$ for training the student model could be written} as:  \squeezeup   
\begin{equation}
L_{T-S} = L_{org} + l_{kd} + \frac{1}{4}\sum_{i=1}^{4}l_{at}^i
\end{equation}
\suiyi{w}here $L_{org}$ is \suiyi{the} distillation-free loss given in equation (\ref{eq:overall}).


 \squeezeup
\section{Experiment}
\label{sec:Ex} \squeezeup
The proposed TGV dataset was employed to benchmark the NR metrics designed for quality assessment of gaming videos as summarized in section~\ref{sec:RW}. The entire dataset was divided into 80\%, 20\% as training and hold-out testing sets respectively. It is worth mentioning that there is no overlap in terms of gaming contents between the training and testing set to ensure the generality of the model. Apart from those NR metrics, 4 commonly used, especially in treaming industry~\cite{ling2020towards}, full reference metrics were also tested, including the Peak Signal to Noise Ratio (PSNR)~\cite{wang2009mean}, the Structural Similarity Index (SSIM)~\cite{wang2004image}, the Multi-Scale Structural Similarity (MS-SSIM)~\cite{wang2003multiscale}, and the Video Multi-Method Assessment Fusion (VMAF)~\cite{li2018vmaf}. To evaluate the performances of the considered objective quality metrics, commonly used performance evacuation methodology including the Pearson correlation coefficient (PCC), the Spearman’s rank order correlation coefficient (SCC), the Kendall correlation coefficient (KCC) and the Root Mean Squared Error (RMSE) are computed between the mean opinion scores and the predicted quality scores. 
 
During fine-tuning (training), all the frames extracted from videos were first rescaled to 1080P with zero paddings. To augment the data, the input frames were further randomly cropped into $540 \times 960$, and flipped left to right or vice versa. The initial learning rate was set as $10^{-04}$, and the Cosine Annealing learning rate decay strategy was applied. We used the Adam optimizer, and the training was stopped after 50 epochs. To avoid overfitting, weight decay was set as $5^{-04}$. $\lambda$ in equation (\ref{eq:overall}) was set to 1.


\textbf{Overall performance:} The results are shown in Table~\ref{tab:res}, where the best performances are highlighted in bold. \suiyi{ERAQUE$_{teacher}$, ERAQUE$_{student}$ and ResNet18 tiny,} are the original teacher backbone model, the student model after distillation and the ResNet18 tiny that is trained from scratch correspondingly. In general, the proposed ResNet18$_{teacher}$ achieves the best performance among all the compared metrics, including the full reference metrics. It even far surpasses the other CNN based state-of-the-art metric NDNetgaming, with PCC values of $0.9646 \ vs.\  0.8579$. The distilled version of our model also outperforms the other metrics. These observations demonstrate the effectiveness of our approach. The performances of the two ResNet18 tiny models also elucidate the advantage of training  light-weight models via knowledge distillation.

\begin{table}[!htbp]
\begin{center}
 \squeezeup  \squeezeup 
\caption{\label{tab:main performance}%
Performances comparison of considered Metrics. }
{
\renewcommand{\baselinestretch}{1}\footnotesize
\begin{tabular}{|c|c|c|c|c|}
\hline
&\bf{PCC} &\bf{SCC} &\bf{KCC} &\bf{RMSE}      \\ \hline
 \multicolumn{5}{ |c| }{ Full Reference Metric (FR)}\\ \hline
PSNR~\cite{wang2009mean} & 0.7467 & 0.7348 & 0.5445 &  0.7077 \\ \hline
SSIM~\cite{wang2004image} & 0.8709 & 0.8654  & 0.6763 & 0.5229 \\ \hline
MS-SSIM~\cite{wang2003multiscale}  &  0.8492 & 0.8457 & 0.6536 & 0.5618\\ \hline 
VMAF~\cite{li2018vmaf} &0.9130 & 0.9102 & 0.7436 &  0.4341 \\ \hline
 \multicolumn{5}{ |c| }{ No Reference Metric (NR)}\\ \hline
GamingPara~\cite{zadtootaghajMMSys}  &  0.5568  & 0.4402 &  0.3268  &  0.9084 \\ \hline
 ITU-T G.1072~\cite{g1072} & 0.0633  & 0.0534 &  0.0412 & 1.0628 \\ \hline
  DEMI~\cite{zadtootaghaj2020demi} &  0.7536 &  0.7455  &  0.5382 &   0.7479 \\ \hline
 NDNetgaming~\cite{NDNetgaming} & 0.8579 &  0.8477 &  0.6578  & 0.5845  \\ \hline
ResNet18 tiny &  0.8298  & 0.8219   &  0.6398   &   2.6031 \\ \hline
\suiyi{ERAQUE$_{student}$}  &  0.9646 &   0.9641 &  0.8436  &    2.6387  \\ \hline
\suiyi{ERAQUE$_{teacher}$} & \bf{0.9714} & \bf{0.9712} & \bf{0.8635}  & \bf{0.2697}   \\ \hline
\end{tabular}}
\end{center}
 \squeezeup  \squeezeup
\label{tab:res}
\end{table}

\textbf{Advantage of knowledge distillation:} Information of the teacher, student networks are summarized in Table~\ref{tab:para}, where $\# para$ indicates the number of network parameters. It is obvious that the complexity of the student network is around half of the teacher model in any respect. To further verify whether the difference of performance between the teacher and student is significant, \textsl{i.e.,} whether the predicted scores are significantly different, the F-test analysis based on the residual difference between the predicted objective scores and the subjective Mean Opinion Score (MOS) values as described in~\cite{ling2019prediction} was employed. The obtained results indicate no significant different between the performances of the proposed model and the distilled model, which demonstrates the efficiency of the teacher-student model distillation strategy.

\begin{table}[!htbp]
\begin{center}
 \squeezeup  \squeezeup
\caption{\label{tab:main performance}%
Comparison of teacher, student models.}
{
\renewcommand{\baselinestretch}{1}\footnotesize
\begin{tabular}{|c|c|c|c|}
\hline
&\bf{$\# para$} &\bf{FLOPs} &\bf{storage size of the model}     \\ \hline
\suiyi{ERAQUE$_{teacher}$}&  11 M &  18.91 G  &  45 M       \\ \hline
\suiyi{ERAQUE$_{student}$} &  5.4 M & 9.28 G  &  21 M     \\ \hline
\end{tabular}}
\end{center}  \squeezeup  \squeezeup
\label{tab:para}
\end{table}

 \squeezeup  \squeezeup
\section{Conclusion}
\label{sec:Con}
In this study, a large-scale subjective study for the quality assessment of mobile gaming videos. According to our best knowledge, our TGV is one of the largest existing relevant dataset. To better quantify the quality of mobile gaming contents, a novel quality assessment framework has also been presented. According to the experimental results, the proposed metrics outperform state-of-the-art quality metrics, and our model distillation strategy could achieve high trade-off between the model complexity and the performance.

\balance 
\bibliographystyle{IEEEbib}
\scriptsize{
\bibliography{refs}}

\end{document}